\documentclass[runningheads]{llncs}
\usepackage{xcolor}
\usepackage{graphicx}
\usepackage{multirow}
\usepackage{tabularx}
\usepackage{tabularx}
\graphicspath{{./figures/}}
\begin{document}

\title{Medical image segmentation with imperfect 3D bounding boxes}
\titlerunning{Segmentation with imperfect 3D bounding boxes}
\author{Ekaterina Redekop, Alexey Chernyavskiy}
%\author{Anonymous}
%

\institute{Philips AI Research, Moscow, Russia
\email{redekop.ep@gmail.com, alexey.chernyavskiy@philips.com}\\}
\maketitle              % typeset the header of the contribution
\begin{abstract}
The development of high quality medical image segmentation algorithms depends on the availability of large datasets with pixel-level labels. The challenges of collecting such datasets, especially in case of 3D volumes, motivate to develop approaches that can learn from other types of labels that are cheap to obtain, e.g. bounding boxes. We focus on 3D medical images with their corresponding 3D bounding boxes which are considered as series of per-slice non-tight 2D bounding boxes. While current weakly-supervised approaches that use 2D bounding boxes as weak labels can be applied to medical image segmentation, we show that their success is limited in cases when the assumption about the tightness of the bounding boxes breaks. We propose a new bounding box correction framework which is trained on a small set of pixel-level annotations to improve the tightness of a larger set of non-tight bounding box annotations. The effectiveness of our solution is demonstrated by evaluating a known weakly-supervised segmentation approach with and without the proposed bounding box correction algorithm. When the tightness is improved by our solution, the results of the weakly-supervised segmentation become much closer to those of the fully-supervised one.
 
\end{abstract}
\section{Introduction}
Automatic solutions for medical image segmentation are designed to increase the work efficiency of medical practitioners, as manual segmentation is an error-prone and time-consuming process. Deep convolutional neural networks (CNN) are known to achieve state-of-the-art performance for this task. However, their success highly depends on the availability of large collections of pixel-level annotations performed by experts. Drawing masks for a 2D image typically requires $\sim 8x$ more time than delineating a bounding box, and $\sim 78x$ more time than assigning an image-level label \cite{bearman2016s}.
For 3D medical images, the need to have high-quality pixel-level labels makes the manual annotation even more tedious. This motivates to develop methods that leverage large amounts of data labelled by weak annotations that are cheaper to obtain. There exist various forms of weak labels, including image-level tags, scribbles, bounding boxes. We focus on the latter ones as they are simple, cheap in terms of annotation cost and, moreover, they provide the localization information about an object. Weak labels can be used individually in the context of weak supervision, or in combination with a small set of accurate pixel-level annotations for semi-supervised learning. 

In the 2D case, bounding boxes can be defined by the coordinates of two opposing corners. In 3D, bounding boxes can be defined either as a series of coordinates of two corners for each slice along a chosen axis, or by three corner coordinates for the entire 3D volume. As we will show, the first alternative is suitable for applying existing weakly- or semi-supervised 2D image segmentation methods. The downside is that the delineation of a bounding box in each 2D layer of the entire volume is time-consuming. The second alternative allows to obtain bounding boxes easily, by quick inspection of a region of interest in three dimensions, but the quality of segmentation approaches can drop increasingly when they are trained using bounding boxes that are far from being tight.

Our contribution is three-fold. First, we show the limitation of current weakly-supervised approaches that use 2D bounding boxes as weak labels, when applied to medical image segmentation in cases when the bounding boxes are not tight. Second, we propose a bounding box correction framework which shrinks the bounding boxes closer to the actual size of the object cross-section in each slice of the 3D volume. Finally, we demonstrate that the proposed solution allows increasing the accuracy of 3D computed tomography (CT) segmentation algorithms trained using pseudo-annotations generated from weak labels.

\section{Related work}
Weakly-supervised learning methods can significantly reduce the cost of annotation that is needed to collect a training set. The methods differ by the type of weak annotation they rely on, such as image-level labels~\cite{wei2018revisiting}, points~\cite{bearman2016s}, partial labels~\cite{xu2015learning} or global image statistics~\cite{bateson2019constrained}.
In this work, we build upon the recent papers that have focused on training neural networks using pseudo-annotations generated from bounding boxes. In \cite{xu2019bounding} and \cite{xu2020new}, Xu et al. formulated the weakly-labeled segmentation as a sparse boundary point detection task solved by training a CNN that predicts the offsets from the given bounding box to the true object boundary. In  \cite{kervadec2020bounding}, Kervadec et al. proposed an image segmentation approach based on global constraints derived from bounding box annotations, including the deep tightness prior and background emptiness constraint. The use of these priors allowed the authors of \cite{kervadec2020bounding} to significantly outperform DeepCut \cite{rajchl2016deepcut} which also relied on bounding boxes for supervision. In \cite{kulharia2020box2seg}, bounding boxes are treated as noisy labels, and per-class attention maps are produced to guide the cross-entropy loss to focus on foreground pixels. 

Semi-supervised learning is the ability of neural networks to derive information from limited sets of labeled data. The authors of \cite{mittal2019semi} proposed an algorithm for semi-supervised semantic image segmentation based on adversarial
training with a feature matching loss to learn from unlabeled images. The approach of Ouali et al.~\cite{ouali2020semi} to the same task is based on cross-consistency, where the general idea of consistency loss is to encourage smooth predictions of the same data under different perturbations.  
Ibrahim et al.~\cite{ibrahim2020semi} proposed to train a primary segmentation model on a small fully-labeled dataset with the aid of an ancillary model that generates segmentation labels for a larger weakly-labeled dataset. In this work, we also use the advantage of a small set of accurately labeled data to train a bounding box correction framework. 

\section{Methodology}

\subsection{Bounding box correction}
\begin{figure}
\centering
\begin{minipage}[b]{.2\linewidth}
  \centering
  \centerline{\includegraphics[width=2.8cm]{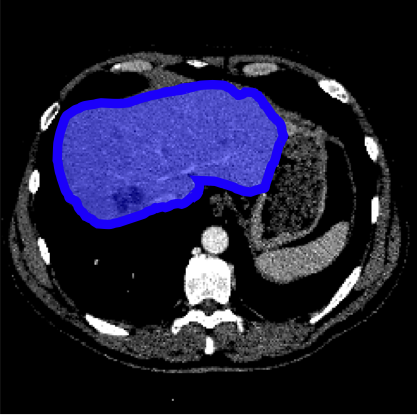}}
%  \vspace{1.5cm}
  \centerline{(a)}\medskip
\end{minipage}
\hfill
\begin{minipage}[b]{0.2\linewidth}
  \centering
  \centerline{\includegraphics[width=2.8cm]{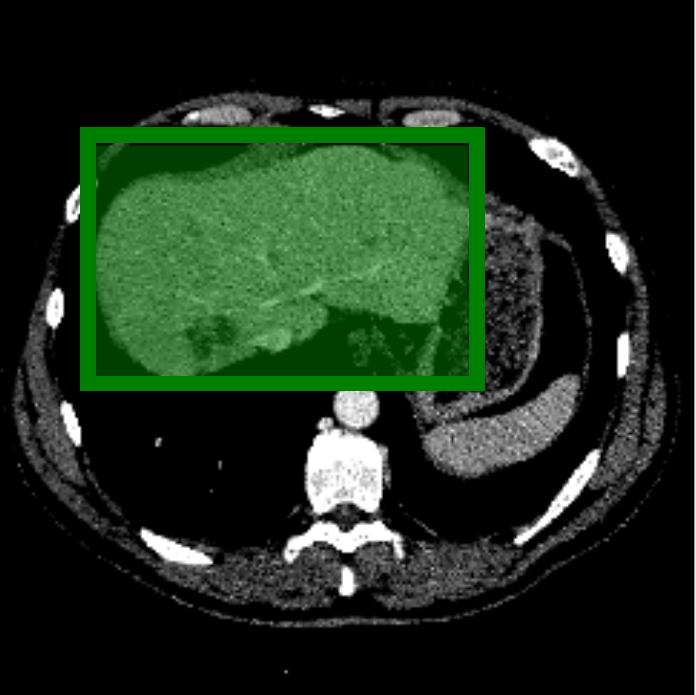}}
%  \vspace{1.5cm}
  \centerline{(b)}\medskip
\end{minipage}
\hfill
\begin{minipage}[b]{0.2\linewidth}
  \centering
  \centerline{\includegraphics[width=2.8cm]{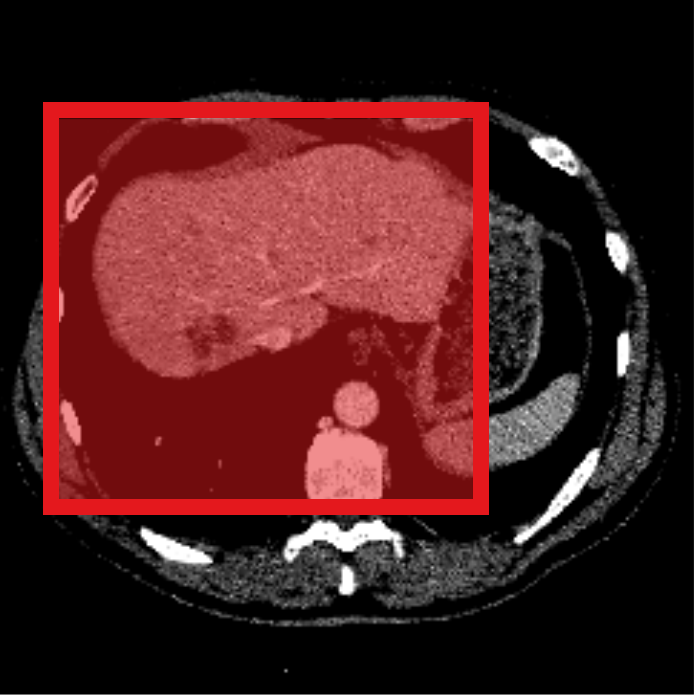}}
%  \vspace{1.5cm}
  \centerline{(c)}\medskip
\end{minipage}
\hfill
\begin{minipage}[b]{0.2\linewidth}
  \centering
  \centerline{\includegraphics[width=2.8cm]{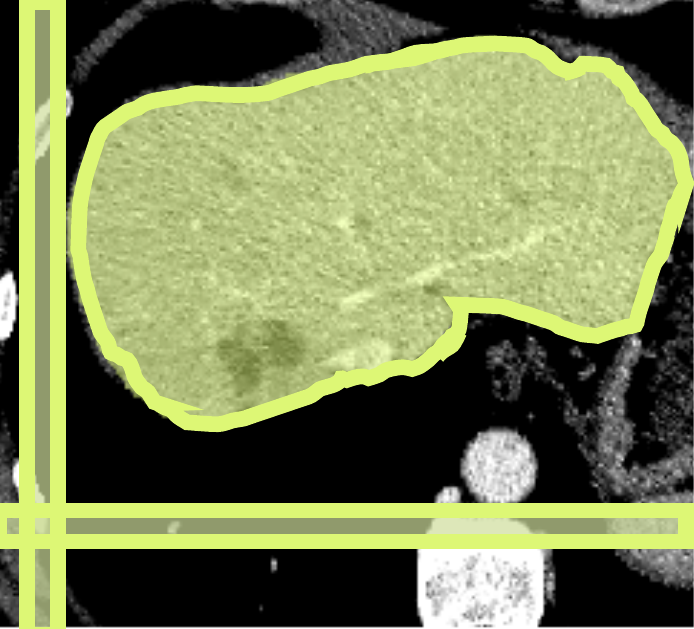}}
%  \vspace{1.5cm}
  \centerline{(d)}\medskip
\end{minipage}
\caption{(a) Ground truth mask, (b) tight bounding box for ground truth mask, (c) non-tight bounding box for a 2D slice of the 3D volume, (d) breaking of the bounding box tightness assumption (see Section 3.2).}
\label{fig_liver_bbox_ex}
\end{figure}

 Consider a three-dimensional object within a volume. It is straightforward to produce a 3D bounding box of the object by finding its extreme points in the three coordinate axes. While this 3D bounding box will be tight in the 3D sense, its rectangular cross-sections will not, in general, remain tight with respect to the planar cross-sections of the volume. Fig.~\ref{fig_liver_bbox_ex} illustrates such a case for the task of liver segmentation in a CT volume. In the Experiments section we show that the success of existing 2D weakly-supervised segmentation methods relies on the bounding boxes being tight and therefore the tightness of the individual 2D bounding boxes should be corrected before training and applying a segmentation CNN. 

% \begin{figure}
% \centering
% \includegraphics[width=0.8\textwidth]{figures/Liver_bbox_example.png}
% \caption{(a) GT mask, (b) tight bounding box for GT mask, (c) non-tight bounding box on 2D slice of 3D volume} \label{fig_liver_bbox_ex}
% \end{figure}
\begin{figure}
\centering
\includegraphics[width=\textwidth]{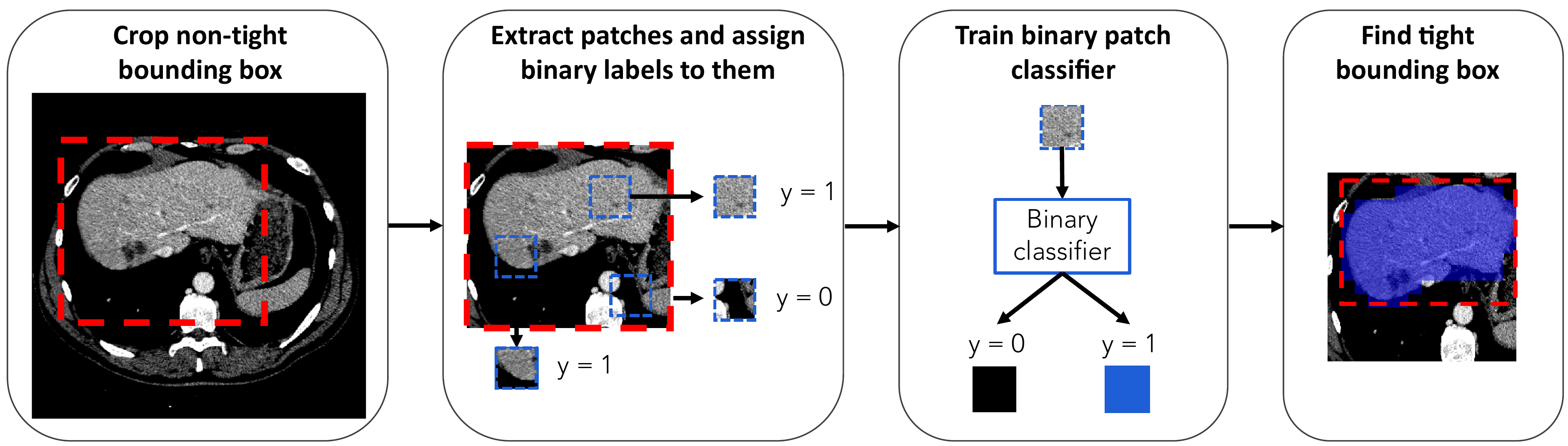}
\caption{Bounding box correction framework.} \label{fig_pipeline_bbox_correction}
\end{figure}

We propose a method to improve the tightness of bounding boxes by using a patch-based classification neural network. The network is trained on a limited subset of ground truth data which is accurately annotated on a pixel level. The proposed solution consists of four steps shown in Fig.~\ref{fig_pipeline_bbox_correction}.  First, a non-tight bounding box is cropped from each slice of the 3D image. Second, each crop is divided into patches of size $p\times p$ pixels, with an overlap equal to $p/2$. Each patch is assigned a binary label $y$: $y=1$ if the foreground object occupies more than $50\%$ of the patch area; otherwise $y=0$. Third, during training and inference, the classification neural network assigns a label for every patch inside the cropped bounding box area. Finally, the patches that the classification network has labeled as foreground determine the extent of the corrected bounding box. During inference, we apply this neural network to data annotated by non-tight bounding boxes, and, following the classification step, obtain more accurate and more tight bounding boxes.  

\subsection{Bounding boxes for weakly supervised segmentation}
We test the bounding box correction method in combination with the novel weakly-supervised framework proposed by Kervadec et al.~\cite{kervadec2020bounding}. The authors perform medical image segmentation by deriving several global constrains from bounding box annotations. In order to regularize the output of the network, they leverage the bounding box tightness prior which was reformulated as a set of global constrains. At the same time, in order to enforce the network to predict no foreground outside the bounding box, the authors add a global background emptiness constraint. The training of a neural network is performed using a sequence of unconstrained losses based on an extension of the log-barrier method. 

The global bounding box tightness prior mentioned above assumes that each of side of the box is sufficiently close to the target region. This means that for any region shape, each vertical or horizontal line inside the bounding box will cross at least one pixel belonging to the target region. This condition does not hold when the provided annotation comes as a 3D bounding box which is represented as a series of per-slice non-tight 2D bounding boxes. In this case, there will exist vertical or horizontal lines shown as stripes in  Fig.~\ref{fig_liver_bbox_ex}~(d), that will lie outside of the actual object boundary. In the Experiments section we demonstrate the poor performance of the weakly-supervised approach from ~\cite{kervadec2020bounding} when the user-provided bounding box is much wider than the true object of interest.

\subsection{Implementation details}
For patch classification that is used for correcting the bounding boxes, we train a VGG-16~\cite{simonyan2014very} CNN using cross-entropy loss. % to classify image patches of size $p=32$ and a shallower VGG-style CNN to classify patches of size $p=16$. 
After the bounding boxes are corrected, we use a residual version of a standard UNet~\cite{kervadec2020bounding} neural network which we trained in a 2.5D manner by taking a stack of three neighboring slices as input and outputting a segmentation for the single central slice of the stack. This approach allows to take advantage of richer spatial information compared to 2D, while requiring less computations compared to 3D CNNs. Following~\cite{kervadec2020bounding}, we trained the segmentation model using the tightness prior in combination with constraint on the global size and masked cross-entropy. We performed three-fold cross-validation to study the variability of image segmentation.

Both the classification and segmentation neural networks are trained using Adam optimizer with learning rate equal to $10^{-4}$. The mini-batch size and the number of epochs are set to 32 and 50 respectively. We set the bounding box tightness prior parameters following~\cite{kervadec2020bounding}. 

\section{Experiments and discussion}
We validate the proposed bounding box correction method, followed by the weakly-supervised segmentation framework, on the liver segmentation dataset provided by the organizers of the Medical Segmentation Decathlon~\cite{simpson2019large}. The data consist of 131 3D contrast-enhanced CT images and was divided into training and validation sets in the proportion 100:31. We normalize the CT data as suggested in~ \cite{isensee2018nnu}. First, the intensity values of pixels that fall under the segmentation masks are collected for the whole training set. Second, the intensity values for the entire dataset are clipped to the [0.5, 99.5] percentiles of the collected values. Third, z-score normalization is applied based on the statistics of the collected values.
 
To train the bounding box correction framework, we further divided the training set into a small subset of accurate pixel-level and a larger subset of weak bounding box annotations, with the size of the small subset equal to 5\%, 10\% and 20\% of the whole training set. We also studied the effect of patch size on the tightness of corrected bounding boxes and on the segmentation accuracy, which was measured as the Dice similarity coefficient between the CNN outputs and ground truth masks.

\subsection{Weakly-supervised segmentation of 3D CT volume using bounding box correction}
In Table~\ref{table_liver_results} we compare the performance of fully- and weakly-supervised training strategies for the liver CT dataset, where 3D voxel-level segmentation masks are available for each 3D CT scan. In this case one can easily obtain a 3D bounding box for the entire object of interest, or a series of tight 2D bounding boxes corresponding to each individual cross-section of the object. The first alternative implies the absence of the bounding box tightness property on most of the 2D slices of the volume.

\begin{figure}
\centering
\begin{minipage}[b]{.2\linewidth}
  \centering
  \centerline{\includegraphics[width=2.8cm]{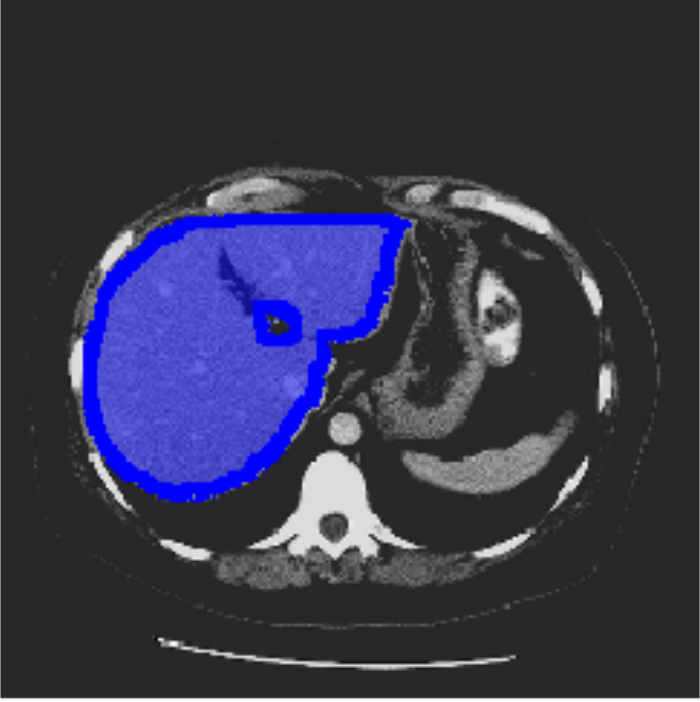}}
%  \vspace{1.5cm}
  \centerline{(a)}\medskip
\end{minipage}
\hfill
\begin{minipage}[b]{0.2\linewidth}
  \centering
  \centerline{\includegraphics[width=2.8cm]{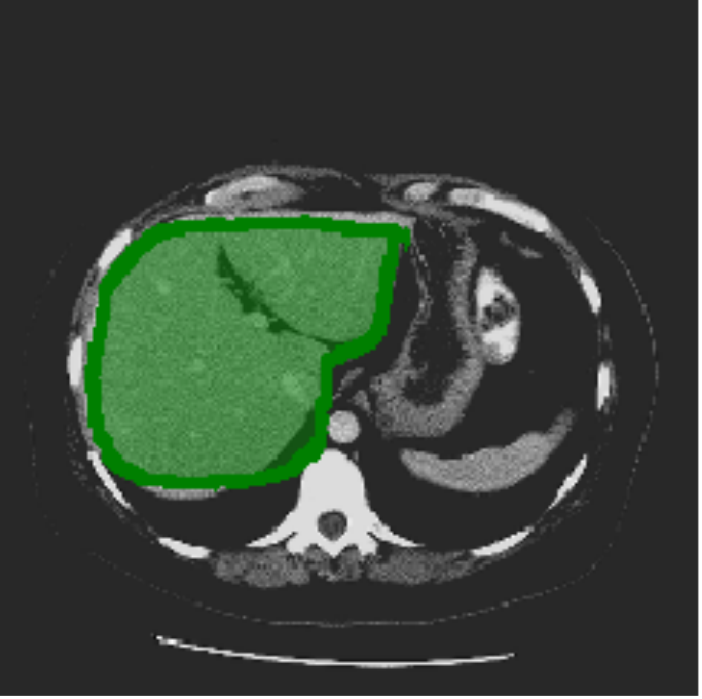}}
%  \vspace{1.5cm}
  \centerline{(b)}\medskip
\end{minipage}
\hfill
\begin{minipage}[b]{0.2\linewidth}
  \centering
  \centerline{\includegraphics[width=2.8cm]{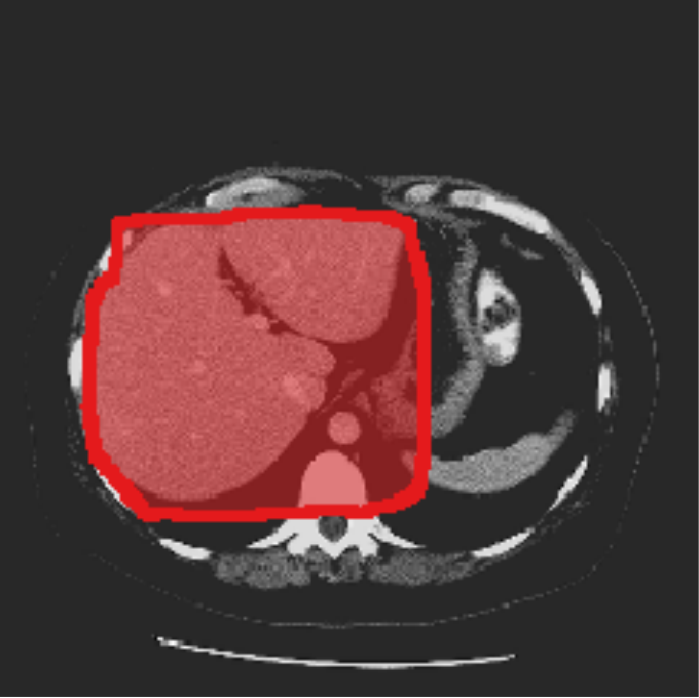}}
%  \vspace{1.5cm}
  \centerline{(c)}\medskip
\end{minipage}
\hfill
\begin{minipage}[b]{0.2\linewidth}
  \centering
  \centerline{\includegraphics[width=2.8cm]{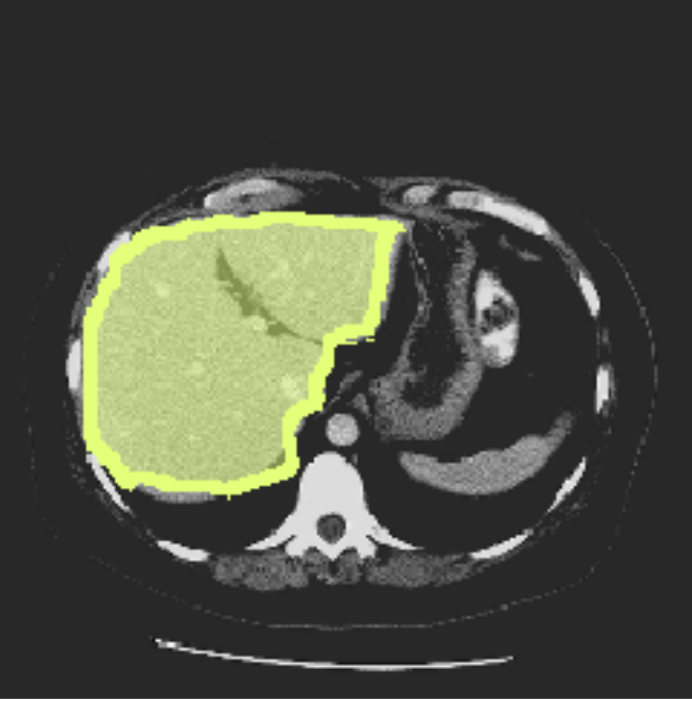}}
%  \vspace{1.5cm}
  \centerline{(d)}\medskip
\end{minipage}
\caption{(a) Ground truth mask. Segmentation results of a 2.5D UNet trained on: (b) 2D tight bounding boxes, (c) 3D non-tight bounding boxes, (d) 3D corrected bounding boxes.}
\label{fig_tightness_res}
\end{figure}

The weakly-supervised approach ~\cite{kervadec2020bounding} can be efficiently applied to slice-wise 2D image segmentation of a 3D object, as long as the ground truth labels are given as tight bounding boxes for each image slice (Table~\ref{table_liver_results}). %We call this experiment 'Weak supervision 2D' and show its results on the validation set in Table~\ref{table_liver_results}. 
If, instead of a series of 2D bounding boxes, the ground truth segmentation labels are given as a 3D bounding box computed over the entire object of interest embedded within a 3D image, then, depending on the shape of the object, the edges of many rectangular 2D cross-sections of a 3D bounding box will be quite distant from the boundaries of the object. In this case, the performance of the semi-supervised approach drops considerably. In order to boost the performance of segmentation networks trained on this kind of weak and noisy labels, we use the advantage of the proposed bounding box correction framework described in section 3.1 and pictured in Fig.~\ref{fig_pipeline_bbox_correction}. In Table~\ref{table_iou} we show the bounding box tightness computed as the intersection over union (IoU) between the tight bounding box generated from a 2D ground truth mask, and the bounding box coming from a 3D box before and after applying our correction procedure on the liver CT dataset. As shown in Table~\ref{table_liver_results} ('3D corrected'), the improvement of bounding box tightness using our approach results in higher segmentation accuracy of models trained with weak supervision. The experiments also show that a smaller patch size ($p=16$) used for correcting the bounding boxes results in higher segmentation accuracy. The amount of accurately labeled data that is used to train the bounding box correction network also plays the role in the final segmentation accuracy. By providing 20 examples one can achieve the quality that is comparable to the performance of segmentation models trained using tight 2D bounding boxes.
\begin{figure}
\centering
\begin{minipage}[b]{.2\linewidth}
  \centering
  \centerline{\includegraphics[width=2.8cm]{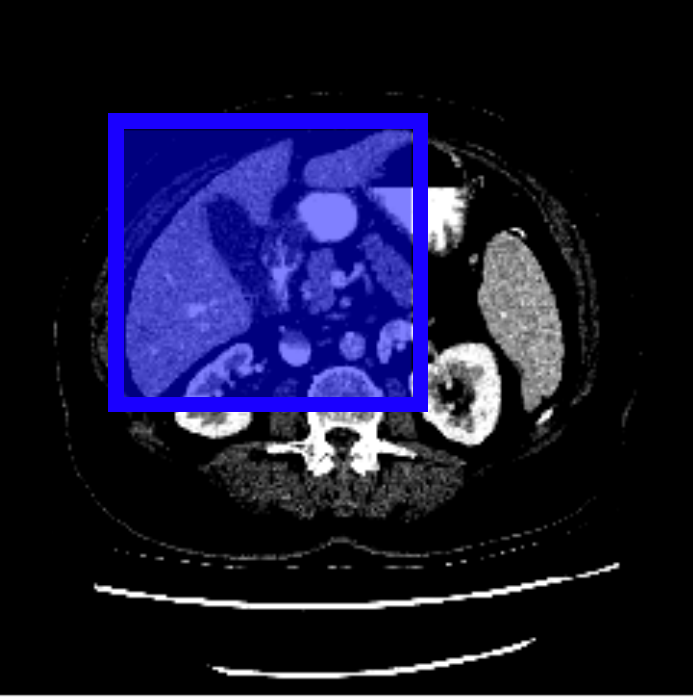}}
%  \vspace{1.5cm}
  \centerline{(a)}\medskip
\end{minipage}
\hfill
\begin{minipage}[b]{0.2\linewidth}
  \centering
  \centerline{\includegraphics[width=2.8cm]{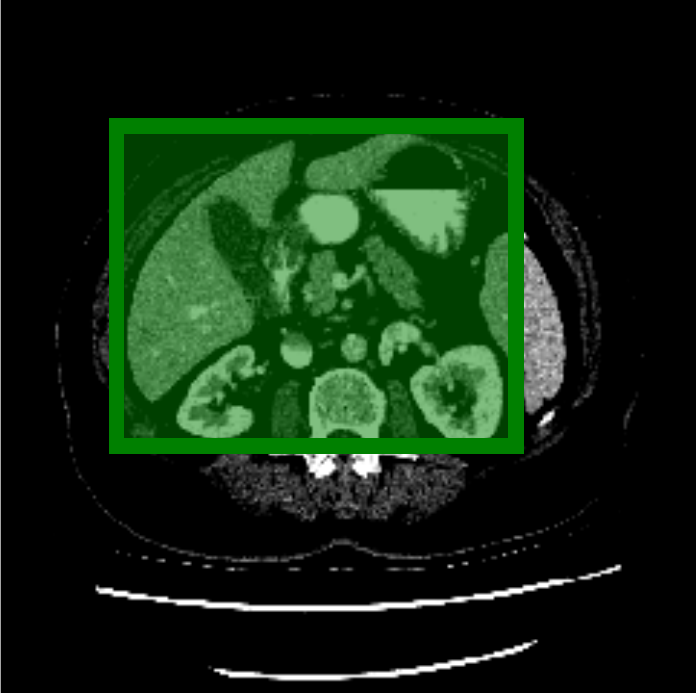}}
%  \vspace{1.5cm}
  \centerline{(b)}\medskip
\end{minipage}
\hfill
\begin{minipage}[b]{0.2\linewidth}
  \centering
  \centerline{\includegraphics[width=2.8cm]{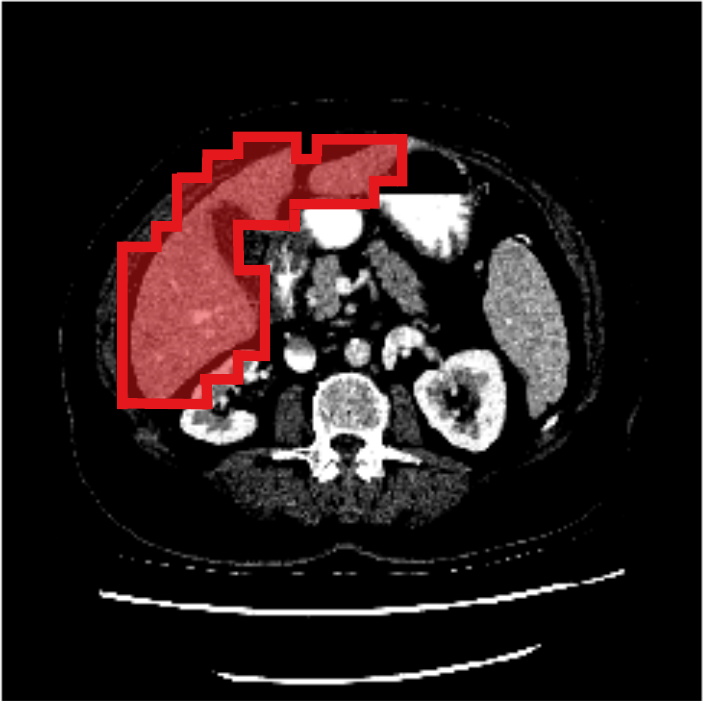}}
%  \vspace{1.5cm}
  \centerline{(c)}\medskip
\end{minipage}
\hfill
\begin{minipage}[b]{0.2\linewidth}
  \centering
  \centerline{\includegraphics[width=2.8cm]{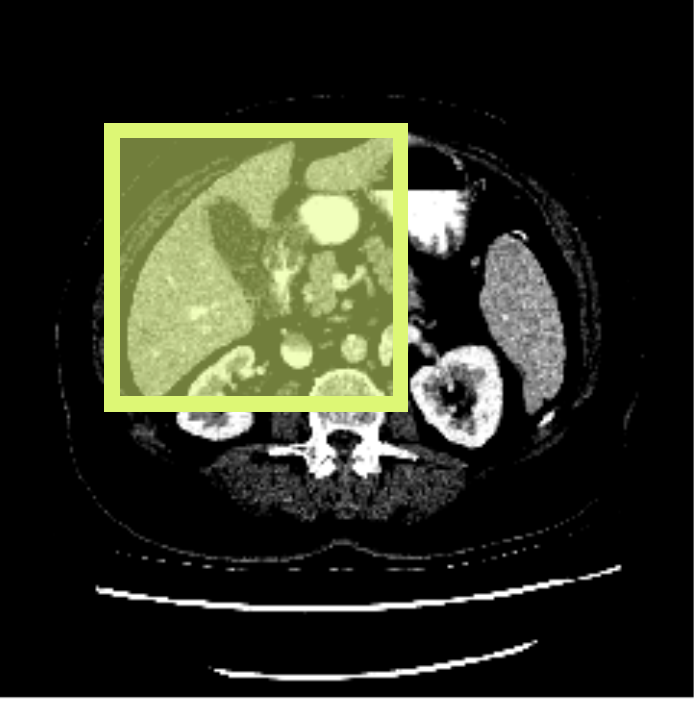}}
%  \vspace{1.5cm}
  \centerline{(d)}\medskip
\end{minipage}
\caption{(a) Tight bounding box for ground truth mask, (b) non-tight bounding box for a 2D slice of the 3D volume, (c) output of patch classification CNN, (d) corrected bounding box for a 2D slice of the 3D volume.}
\label{fig_bbox_ref_res}
\end{figure}
\begin{table}
\caption{Segmentation of 3D liver CT images: Dice scores obtained by training segmentation models using weak supervision provided as tight, non-tight and corrected bounding boxes. The Dice score for full supervision (voxel-level masks) is 0.92$\pm$0.03.}\label{table_liver_results}
\centering
%\begin{tabular}{lccccc}
%Type of weak supervision                      & 2D        &  3D         & \multicolumn{3}{c}{ 3D corrected}  \\ \hline
%\ & \ & \ & \multicolumn{3}{c}{\# images for box correction} \\ 
%\ & \ & \ &  5         & 10         & 20                  \\ \hline
%Patch size $p=16$                       & 0.90$\pm$0.01 & 0.28$\pm$0.06 & 0.83$\pm$0.02 & 0.86$\pm$0.01 & 0.90$\pm$0.04\\ \hline
%Patch size $p=32$                       & 0.90$\pm$0.01 & 0.28$\pm$0.06 & 0.82$\pm$0.04 & 0.84$\pm$0.03 & 0.89$\pm$0.02\\ \hline
%\end{tabular}
%\begin{tabular}{cccccc}
% \begin{tabularx}{350pt}{XXXXXX} 
\begin{tabularx}{345pt}{*6{>{\centering\arraybackslash}X}}
\multirow{2}{*}{} & \multirow{2}{*}{} & & \multicolumn{3}{c}{3D corrected}  \\ 
\ & \ & \ & \multicolumn{3}{c}{number of images for box correction}  \\
2D tight & 3D non-tight & & 5 & 10 & 20 \\ \hline
\multirow{2}{*}{0.90$\pm$0.01} & \multirow{2}{*}{0.28$\pm$0.06} & $p=16$ & 0.83$\pm$0.02 & 0.86$\pm$0.01 & \textbf{0.90$\pm$0.04}\\
& & $p=32$ & 0.82$\pm$0.04 & 0.84$\pm$0.03 & 0.89$\pm$0.02\\ \hline
\end{tabularx}
\end{table}

\begin{table}
\centering
\caption{Mean IoU values before and after 3D bounding box correction computed with respect to tight slice-wise ground truth bounding boxes. \label{table_iou}}
%\begin{tabular}{lccc}
%\#images for box correction    & 5   & 10   & 20\\ 
%\hline
%\ 3D bounding box         & 0.15$\pm$0.01 & 0.15$\pm$0.01 & 0.17$\pm$0.02    \\
%\ Patch size $p=16$ & 0.87$\pm$0.02 & 0.89$\pm$0.04 & 0.93$\pm$0.04  \\ 
%\ Patch size $p=32$ & 0.89$\pm$0.01 & 0.89$\pm$0.02 & 0.92$\pm$0.01  \\ 
%\hline
%\end{tabular}

\begin{tabularx}{345pt}{*6{>{\centering\arraybackslash}X}}
\multirow{3}{3cm}{\centering 3D non-tight bounding box} & & \multicolumn{3}{c}{3D corrected}  
  \\ 
&  & \multicolumn{3}{c}{number of images for box correction}  \\
&  & 5 & 10 & 20 \\ \hline
\multirow{2}{*}{0.15$\pm$0.01} & $p=16$ & 0.87$\pm$0.02 & 0.89$\pm$0.04 & \textbf{0.93$\pm$0.04}\\
& $p=32$ & 0.89$\pm$0.01 & 0.89$\pm$0.02 & 0.92$\pm$0.01\\ \hline
\end{tabularx}
\end{table}

\section{Conclusions and Discussions}
We have addressed the main limitation of a known approach to weakly-supervised 2D and 3D medical segmentation that assumes that the labels, coming in the form of two-dimensional bounding boxes, are tight. We have shown that in a practical case when a single 3D bounding box is provided for the whole object, the tightness of 2D slice-wise bounding boxes deteriorates, which results in poor segmentation accuracy of neural networks trained with this type of supervision. We have proposed a bounding box correction framework that improves the tightness by using a a patch-based classification network trained on a small subset of pixel-level annotated data. By producing higher quality annotations out of weak labels, our approach allows to increase the accuracy of 3D medical weakly-supervised segmentation. 

Since the performance of the proposed approach may depend on the soft tissue contrast, its applicability for segmentation of other organs is yet to be investigated. The patch size and the share of fully-supervised samples used for bounding box correction may play an important role.

\bibliographystyle{splncs04}
\bibliography{refs}





% LNCS Homepage, \url{http://www.springer.com/lncs}. Last accessed 4
% \end{thebibliography}
\end{document}